\documentclass{article}
\PassOptionsToPackage{numbers,compress}{natbib}

\usepackage[preprint]{neurips_2026}
\usepackage{graphicx}
\usepackage{caption}
\usepackage{amssymb}
\usepackage[dvipsnames]{xcolor}
\usepackage{colortbl}

\definecolor{maroon}{cmyk}{0,0.87,0.68,0.32}

\usepackage{amsmath}
\usepackage[utf8]{inputenc}
\usepackage[T1]{fontenc}
\usepackage{hyperref}
\usepackage{url}
\usepackage{booktabs}
\usepackage{amsfonts}
\usepackage{nicefrac}
\usepackage{microtype}
\usepackage{xcolor}

\title{GS-VLA: Plug-and-Play
Viewpoint Canonicalization \\ for Frozen VLA Policies via Gaussian Splatting }

\author{
  Yechan Park\\
  Dankook University\\
  Yongin, Republic of Korea\\
  \texttt{yechan3219@dankook.ac.kr}\\
  \And
  HyunJin Kim\\
  Dept. of EEE, Dankook University\\
  Yongin, Republic of Korea\\
  \texttt{hyunjin2.kim@gmail.com}\\
  \emph{(Corresponding author)}\\
  }

\begin{document}

\maketitle

\begin{figure}[h]
    \centering
    \includegraphics[width=1.0\textwidth]{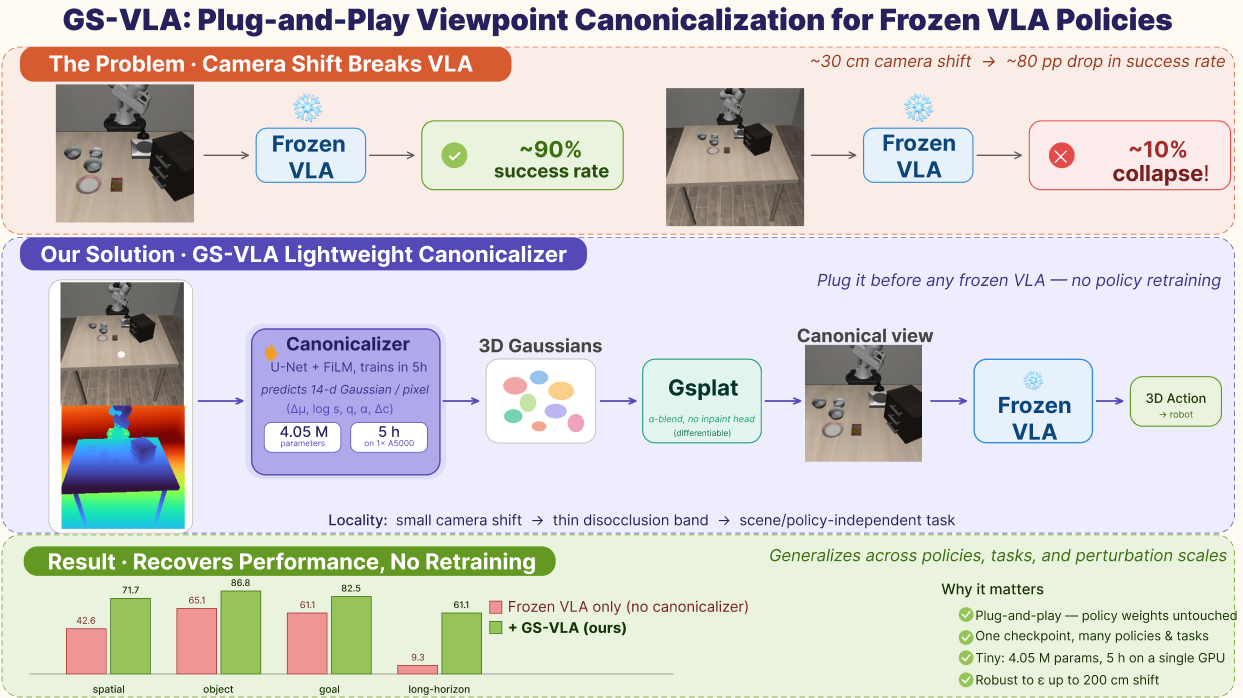}
    \label{fig:overview}
\end{figure}

\begin{abstract}

This paper proposes a lightweight, plug-and-play framework that improves robustness to viewpoint shifts in Vision-Language-Action (VLA) policies without policy retraining. To our knowledge, this is the first approach to directly leverage 3D Gaussian-based novel-view synthesis for observation-space adaptation in VLA policies.
Current VLA performance relies on the implicit assumption that training and deployment camera configurations are identical. 
Our experiments show that even a small displacement of the camera mount can reduce the success rate on the LIBERO benchmark from about 90\% to about 10\% in the worst case.
Prior approaches, such as large-scale fine-tuning or generative data augmentation, are computationally expensive and risk catastrophic forgetting. To address this, viewpoint shifts are reformulated as a localized novel-view synthesis problem. Under a \textit{Locality} assumption, that camera perturbations remain within a small bounded region relative to the workspace, viewpoint normalization reduces to a scene- and policy-independent disocclusion task. Our work implements this idea with a 4M-parameter 3D-Gaussian canonicalizer prepended to a frozen VLA policy. Without modifying policy weights, GS-VLA improves performance across three orthogonal axes: (1) \textbf{Policy architectures}, (2) \textbf{Unseen task suites}, and (3) \textbf{Perturbation scales}. These results show that a lightweight visual module can recover a large fraction of the performance lost under viewpoint shift, without policy retraining.

\end{abstract}

\section{Introduction}
\label{intro}

Vision-Language-Action (VLA) models~\cite{rt1,rt2,octo,openvla,openvla-oft,pi0,pi05,dexvla,DP} have become a standard policy class for general manipulation. They perform strongly on benchmarks such as LIBERO~\cite{libero} and CALVIN~\cite{calvin} and can transfer to real-world robots without additional modification. 
However, these successes often rely on an implicit assumption: the visual conditions at deployment closely match those used during training. In practice, even small deviations in camera viewpoint can lead to substantial performance degradation, highlighting viewpoint robustness as an important open challenge for reliable real-world deployment.
However, improving robustness through additional data collection and policy fine-tuning is often impractical in real deployment settings, where new demonstrations are costly or unavailable. Also, there is a risk of policy collapse; trying to fine-tune the model for new camera views might ruin its original manipulation skills—a problem often called catastrophic forgetting.

More fundamentally, performance drops under viewpoint shifts do not necessarily reflect failures in high-level reasoning or control capabilities; rather, they may stem from a mismatch between deployment-time observations and those encountered during training. This raises a broader question: \emph{should robustness to viewpoint variation be learned entirely within policy parameters, or can it be achieved by adapting the observation space instead?}

The latter observation space adaptation is not only possible but also cost-effective. 
In many real-world setups, camera changes are usually small and stay within a limited range. In practice, these perturbations are often small relative to the workspace, typically amounting to only a few to a few tens of centimeters. We refer to this practical property as \textit{Locality}.
Under this assumption, viewpoint changes affect only a limited portion of the image, while most pixels remain geometrically recoverable. Specifically, most of the visual content can be mapped to the canonical view using depth information and camera parameters. This means that learning is mainly needed for a narrow missing region, which can be treated as an inpainting problem. Importantly, this problem depends more on the visual domain than on the particular scene or policy. In this sense, viewpoint canonicalization is not a full novel-view-synthesis problem, but a simpler local adaptation problem that is largely independent of the scene and policy. We treat Locality as an empirically testable assumption.

This makes viewpoint adaptation much more practical. 
Under Locality, there is no need for a large view-synthesis model designed for arbitrary scenes and camera baselines.
Instead, a lightweight canonicalization module can be placed before a frozen VLA policy to restore aligned observations. 
This plug-and-play design recovers a substantial portion of the lost performance across different policy architectures, task suites, and perturbation scales without any policy retraining.

The main contributions of this paper are as follows:
\begin{itemize}
\item \textbf{A new formulation of viewpoint adaptation for VLA policies: } 
We show that viewpoint adaptation for frozen VLA policies can be addressed in observation space rather than in policy parameters, and formalize this perspective through \textit{Locality}, 
a practical premise that reduces the problem to local disocclusion handling.

\item \textbf{The first Gaussian-based canonicalization module for VLA policies: } 
To the best of our knowledge, this is the first work to directly use a 3D Gaussian-based module for observation-space adaptation in VLA policies, thereby enabling efficient viewpoint canonicalization without policy retraining. 
The resulting module is lightweight and efficient to train, 
consisting of a 4M-parameter U-Net~\cite{unet} adapter that predicts 14-dimensional Gaussian attributes and converges in only 5 hours on a single RTX A5000.

\item \textbf{Generalization across policies, tasks, and perturbation scales: } 
We demonstrate that a single checkpoint transfers across multiple policy architectures, 
unseen task suites, and large camera perturbations, 
recovering substantial performance without axis-specific re-training.

\item \textbf{Empirical analysis of the recovery mechanism:} 
We provide empirical evidence that most of the target view is handled by deterministic geometric mapping, while learning is needed only for a spatially localized residual region.

\end{itemize} 

Taken together, the above contributions suggest that viewpoint robustness 
for frozen VLA policies can be addressed effectively through lightweight observation-space adaptation, without modifying the policy itself. 
The rest of the paper is organized as follows. 
Firstly, prior works on VLA robustness and novel-view synthesis are reviewed. 
We then present our Locality-based formulation and the proposed 3D Gaussian canonicalization module. Finally, we evaluate the method across multiple policies, task suites, and perturbation scales, and provide an analysis of the recovery mechanism.

\section{Related Works}
\label{gen_inst}

\textbf{VLA Foundation Policies.} Recent VLA models~\cite{rt1,rt2,pi0,pi05,openvla,openvla-oft,spatialvla,dexvla,DP,octo,xvla,act,rdt,ecot,diff3d} unified multi-modal inputs under large Transformer backbones trained on massive cross-embodiment datasets. 
Although these models achieved $\geq 90\%$ success on LIBERO~\cite{libero}, 
their performances were predicated on static training camera setups. 
Unlike works that modified internal activations, 
we treat these multi-billion-parameter models as frozen, black-box policies 
and improve viewpoint robustness entirely in the observation space. 
This preserves the integrity of the policy while avoiding the costs of retraining.

\textbf{Camera-Robust Robot Learning.} 
Standard approaches to viewpoint robustness relied on retraining policies under augmented camera distributions. 
SPARTN~\cite{spartn} utilized NeRF~\cite{nerf} to generate novel-view trajectories for behavior cloning. 
The concurrent AnyCamVLA~\cite{anycamvla} similarly froze the policy but relied on LVSM~\cite{lvsm} synthesizer with dual-view inputs, with evaluation limited to narrow perturbations. 
On the other hand, we introduce a Locality-based reformulation that reduces the number of parameters by 42$\times$ 
compared with AnyCamVLA.

\textbf{Feed-Forward Novel-View Synthesis.} 
Novel-view synthesis (NVS) generates images from new camera poses given one or more input views.
Per-scene radiance fields~\cite{nerf,instant,mipnerf} achieve high fidelity but require per-scene optimization, while feed-forward predictors~\cite{pixelnerf,ibrnet} generalize across scenes.
More recently, diffusion-based generators~\cite{zeronvs,reconfusion} synthesize plausible views even under large baselines, at the cost of tens to hundreds of millions of parameters.
Our contribution lies not in proposing a new NVS architecture but in reformulating the problem itself: by restricting the task to a bounded $\varepsilon$-ball under the \textit{Locality} premise, we reduce capacity requirements by one to two orders of magnitude.

\textbf{3D Gaussian Splatting Variants.}
Building on 3DGS~\cite{3dgs} and its differentiable rasterizer~\cite{gsplat}, recent feed-forward variants predict per-pixel Gaussians from one or more views~\cite{splatter,flash3d,pixelnerf,mvsplat} and improve rendering quality and structure~\cite{mipsplat,scaffold}. Flash3D in particular leverages monocular depth priors but targets cross-scene generalization.Our canonicalizer follows the pixel-aligned line but specializes the task to in-distribution camera perturbations, so the U-Net only needs to shape per-pixel Gaussians within an $\varepsilon$-ball around the training pose rather than synthesize arbitrary views.

\textbf{Image Inpainting.}
Filling disocclusions resembles image inpainting~\cite{lama,ldm}, but state-of-the-art inpainters require tens of millions of parameters to hallucinate semantic content over free-form holes.
In our setting, $\varepsilon$-bounded camera shifts produce only O($\varepsilon$)-thin holes along depth discontinuities, so no hallucination is needed and a 4M-parameter U-Net suffices 

\textbf{VLA Robustness Benchmarks.}
LIBERO-Plus~\cite{libero-plus} recently showed that camera viewpoint is a major source of fragility for state-of-the-art VLAs.
In response to this challenge, we introduce an observation-space remedy and evaluate its effectiveness across three axes: policy, task suite, and perturbation scale.
Unlike concurrent approaches targeting other robustness dimensions, such as SpatialVLA~\cite{spatialvla}, our observation-space remedy is orthogonal to internal policy modifications and can be combined with them to achieve more comprehensive robustness.

\section{Proposed method}
\label{sec:proposed}

This section presents our formulation of the Locality assumption as a practical module. 
We first define the objective of the canonicalizer, then show that Locality allows a computationally expensive novel-view synthesizer to be replaced with a lightweight U-Net. Additionally, the training procedure and the module deployment are described.

\subsection{Motivation}
\label{ssec:motivation}

A VLA policy trained at a fixed camera pose tends to collapse once the camera moves at deployment; in our worst LIBERO suite, the success rate decreases from $92.4\%$ to $9.3\%$. 
A natural remedy is to preprocess the observation back to the canonical view before feeding it to the policy. 
Although it seems like we need to solve a complex NVS problem, the changes in the camera position are actually very small. 
In practical settings, rig-mounted cameras typically drift by only a few centimeters, such that the majority of pixels remain visible after the pose change, 
while only a narrow region near depth discontinuities becomes newly exposed.

That band has area $O(\rho)$ in the perturbation magnitude $\rho$. A geometric argument for this area bound and the underlying geometry are detailed in Appendix \ref{appx:locality}. Therefore, the core challenge of the learning problem reduces to inpainting a thin, newly disoccluded band along depth-discontinuity curves rather than synthesizing a general novel view.
This makes our canonicalizer a natural fit: rather than generating the canonical image from scratch, it geometrically transports the visible source content into the canonical frame, leaving learning to handle only the narrow disoccluded regions. As a result, the learnable component remains highly lightweight—a $4$M-parameter U-Net is all we need, which is over $25\times$ smaller than generic feed-forward NVS predictors~\cite{lvsm,flash3d,splatter}. We embed this same locality structure inside the architecture: each source pixel is back-projected through its depth to anchor a Gaussian primitive, the U-Net shapes each primitive (depth-bounded center offset, anisotropic scale, orientation, opacity, and color residual), and differentiable $\alpha$-splatting~\cite{gsplat} composes them at the canonical pose, jointly resolving occlusion ordering and blending across the disoccluded bands. Architectural details and the 14-dim descriptor are provided in Section \ref{sec:arch}.

\subsection{Problem definition}
\label{Problem definition} 

\paragraph{Setup and notation.} We first define the input-output interface of the viewpoint canonicalizer $f_\phi$. A frozen VLA policy $\pi_\theta$ is trained at a canonical pose (the camera pose seen during $\pi_\theta$ training), $C^\star = (R^\star, t^\star) \in \mathrm{SE}(3)$. Here, $\mathrm{SE}(3)$ denotes the group of 3D rigid transformations, represented by a rotation and a translation without scale, and $\mathrm{SO}(3)$ denotes the group of 3D rotation matrices. We write $R^\star \in \mathrm{SO}(3)$ for the camera rotation and $t^\star \in \mathbb{R}^3$ for its world-space position. 
The camera intrinsics are given by the pinhole matrix $K \in \mathbb{R}^{3 \times 3}$, which encodes the focal lengths and principal point.
At deployment, the camera may move to a different \emph{source pose} $C_s = (R_s, t_s)$. For each frame, the canonicalizer receives an RGB image $I_s \in [0,1]^{H \times W \times 3}$, a per-pixel metric depth map $D_s \in \mathbb{R}_{>0}^{H \times W}$ aligned with $I_s$, and the camera parameters $(C_s, C^\star, K)$. Here, $D_s(u,v)$ denotes the depth, in meters, of the 3D point observed at pixel $(u,v)$ in $I_s$.
Throughout the paper, we use the camera-to-world convention: a 3D point $X_c \in \mathbb{R}^3$ expressed in the camera frame is mapped to the world frame by $X_w = R X_c + t$. The goal of the canonicalizer is to predict the image that would have been observed from the canonical pose, which is formulated as: 
\begin{equation}
f_\phi : (I_s, D_s, C_s, C^\star, K) \longmapsto \widehat{I}^\star \in [0,1]^{H \times W \times 3},
\label{eq:fphi}
\end{equation}
where $\widehat{I}^\star$ is the predicted canonical-view image. The frozen policy then acts on this canonicalized observation, i.e., $a = \pi_\theta(\widehat{I}^\star)$. The policy parameters $\theta$ are never updated, and $f_\phi$ operates strictly on a per-frame basis, without temporal aggregation, multi-view input, or per-scene optimization.

\paragraph{Locality assumption.} 
We formalize the Locality assumption described in \textbf{Introduction} Section.
The deployment pose is assumed to remain within a bounded neighborhood of the canonical pose, with translation tolerance $\varepsilon_t > 0$ and rotation tolerance $\varepsilon_R > 0$.
The Locality set, $\mathcal{B}(C^\star; \varepsilon_t, \varepsilon_R)$, is defined as: 
\begin{equation}
\mathcal{B}(C^\star; \varepsilon_t, \varepsilon_R)
:=
\Bigl\{(R,t) : \lVert t - t^\star \rVert_2 \le \varepsilon_t,\; \lVert \log\!\bigl((R^\star)^\top R\bigr) \rVert_2 \le \varepsilon_R\Bigr\},
\label{eq:locality-ball}
\end{equation}
where 
\begin{equation}
C_s \in \mathcal{B}(C^\star; \varepsilon_t, \varepsilon_R).
\end{equation}
In~\eqref{eq:locality-ball}, the term $\lVert \log\!\bigl((R^\star)^\top R\bigr) \rVert_2$ represents the geodesic angle between the current rotation $R$ and the canonical rotation $R^\star$.
In our experiments, we relate the rotation limit $\varepsilon_R$ to the translation limit $\varepsilon_t$ via a first-order scaling derived from the motion-field equation under a pinhole camera. 
Whereas translation-induced image motion scales inversely with depth, rotation induces depth-independent motion~\cite{interpretation,multi-viewCV}. 
Therefore, we use the approximation $\varepsilon_R \approx \varepsilon_t / d_{\min}$, where $d_{\min}$ denotes the minimum scene depth. 
This allows us to control the overall camera perturbation using a single parameter, $\varepsilon_t$. 
To measure the total pose change, we define a normalized pose distance, $\rho$, as: 
\begin{equation}
 \rho = \frac{\|t_s - t^\star\|}{d_{\min}} + \|\log(R^{\star\top} R_s)\|,
\end{equation}
which combines translation and rotation into a unified scalar. 

Under our Locality assumption, the fraction of newly exposed regions, $\eta(\rho)$, grows linearly to leading order as $\eta(\rho) = \alpha_{\mathrm{par}}\,\rho_{\mathrm{par}} + \alpha_{\mathrm{fr}}\,\rho_{\mathrm{fr}} + O(\rho^2)$ for $\rho \le 1$ in matched units, with $\rho_{\mathrm{par}} = \|t_s - t^\star\|/d_{\min}$ and $\rho_{\mathrm{fr}} = \|\log(R^{\star\top}R_s)\|$.
The two leading coefficients depend only on scene geometry: $\alpha_{\mathrm{par}}$ on the silhouette length $L^\star_{\mathrm{px}}$ and the relative depth contrast $(d_{\max} - d_{\min})/d_{\max}$, and $\alpha_{\mathrm{fr}}$ on the image-frame perimeter. We give a geometric argument for this bound, together with the explicit forms of $\alpha_{\mathrm{par}}$ and $\alpha_{\mathrm{fr}}$, in Appendix~\ref{appx:locality}.

We test this scaling empirically along three additional settings: Section~\ref{sec:eps-sweep} sweeps the perturbation scale $\varepsilon_t \in \{5, 15, 30, 50, 100, 200\}\,\text{cm}$; Section~\ref{sec:extrinsic} stresses calibration noise; and Appendix~\ref{appx:locality-validation} verifies the linearity of $\eta(\rho)$ directly.

\subsection{Pixel-aligned 3D-Gaussian architecture}
\label{sec:arch}

The canonicalizer $f_\phi$ is realized as follows.
Under the Locality assumption, we only need to handle a thin disocclusion band, so we use a compact U-Net rather than a generic NVS backbone~\cite{3dgs,flash3d,splatter}, 
which can reduce computational costs. 
A symmetric U-Net takes the channel-wise concatenation $[I_s \,\|\, D_s]$ as input and is FiLM-modulated~\cite{film} by a pose-pair embedding, which is formulated as $z_{\mathrm{pose}} = \mathrm{MLP}(C_s, C^\star) \in \mathbb{R}^{128}$. 
At every source pixel, the decoder predicts a $14$-dimensional Gaussian descriptor consisting of a position residual ($\Delta\mu$), an anisotropic log-scale ($\log s$), an orientation quaternion ($q$), an opacity ($\alpha$), and a color residual ($\Delta c$). 
All heads are zero-initialized so that $f_\phi$ starts exactly at the geometric warp. 
The Gaussian center is anchored at the world-space back-projection $X_w = R_s K^{-1} D_s(u,v) [u,v,1]^\top + t_s$ and is allowed only a depth-bounded residual,
\begin{equation}
    \mu \;=\; X_w \;+\; c_{\Delta\mu}\,D_s(u,v)\,\tanh(\Delta\mu),
    \qquad c_{\Delta\mu} = 0.1,
    \label{eq:mu}
\end{equation}
which caps displacement at $10\%$ of local depth. 
This keeps the residual from drifting away from the geometric base, which we found to be important for stability.

One Gaussian per source pixel, with RGB $I_s(u,v)+\Delta c$, are rasterized at the canonical pose $C^\star$ by \texttt{gsplat}'s~\cite{gsplat} differentiable $\alpha$-blender, $\widehat{I}^\star(p) = \sum_{i \in \mathcal{N}(p)} T_i \alpha_i c_i$ with $T_i = \prod_{j<i}(1-\alpha_j)$. Occlusion is resolved by depth-sorting, and the $\alpha$-blender itself fills the disocclusion band, so we do not need a separate inpainting head. The canonicalizer only requires $\mathbf{4}$M parameters, roughly 840 $\times$ smaller than finetuning $\pi_{0.5}$~\cite{pi05} and 42 $\times$ smaller than AnyCamVLA~\cite{anycamvla}.

\subsection{Training}
\label{sec:train}
We intentionally avoid specialized training techniques so that the observed gains can be attributed to the Locality reduction rather than to the training method. We train on only one suite so that the remaining suites can serve as unseen test domains for cross-suite evaluation. The objective is the standard 3DGS~\cite{3dgs} $0.8\!:\!0.2$ $L_1$/SSIM reconstruction loss:                                                                                                            
\begin{equation}                                                                                                                                                                                        
      \mathcal{L} = 0.8\|\widehat{I}^\star \!-\! I^\star\|_1 + 0.2\bigl(1\!-\!\mathrm{SSIM}\bigr).                                                                                                        
      \label{eq:loss}                                                                                                                                                                                     
\end{equation}

\section{Experiments}
\label{sec:experiments}
In our experiments, we evaluate GS-VLA in terms of magnitude, generality, and deployment realism: 
we quantify the extent to which a canonicalizer recovers the performance lost under perturbations, relative to an upper bound given by the unperturbed (best-case) performance and a lower bound corresponding to the perturbed setting without canonicalization.
We also evaluate whether a single checkpoint generalizes across different policies, benchmark suites, and perturbation levels $\varepsilon$.
Additionally, we assess whether the performance gains persist under realistic deployment conditions, specifically in the presence of simultaneous translation--rotation drift and extrinsic-pose calibration error.
All experiments use a \emph{single} checkpoint trained once on \texttt{libero\_spatial} at $\varepsilon_t = 100$\,cm; no axis-specific retraining is performed.

\subsection{Experimental setup}
\label{sec:setup}
\paragraph{Benchmark.}
We evaluate on the four LIBERO suites~\cite{libero}—\texttt{libero\_spatial}, \texttt{libero\_object}, \texttt{libero\_goal}, and \texttt{libero\_10}—which cover diverse manipulation scenarios including spatial reasoning, object-centric interactions, and long-horizon tasks. Performance is measured by task success rate over multiple episodes.

\noindent
\begin{minipage}[t]{0.56\linewidth}
\paragraph{Camera randomization.}
We add random Gaussian noise to both the camera's position and its look-at point around the canonical pose. The parameter $\varepsilon_t$ controls the size of this noise. We then adjust the camera's orientation so it directly faces the new look-at point. If a shifted camera falls below the table, we discard it and sample a new one. Our default setting uses $\varepsilon_t \approx 100\,\mathrm{cm}$, and we apply more noise in the sideways and vertical directions (orthogonal to the view) than in the depth direction. To test the model's robustness at different noise levels, we simply scale this default setting. Figure~\ref{fig:camera_randomization} visualizes the resulting camera distribution. Finally, in order to isolate the effects of calibration errors, we inject independent location and rotation noise into the canonicalizer's input. 
This simulates real-world scenarios in which the camera pose assumed by the system deviates slightly from the true pose.
\end{minipage}
\hfill
\begin{minipage}[t]{0.40\linewidth}
\vspace{-1.5em}
\centering
\includegraphics[width=\linewidth]{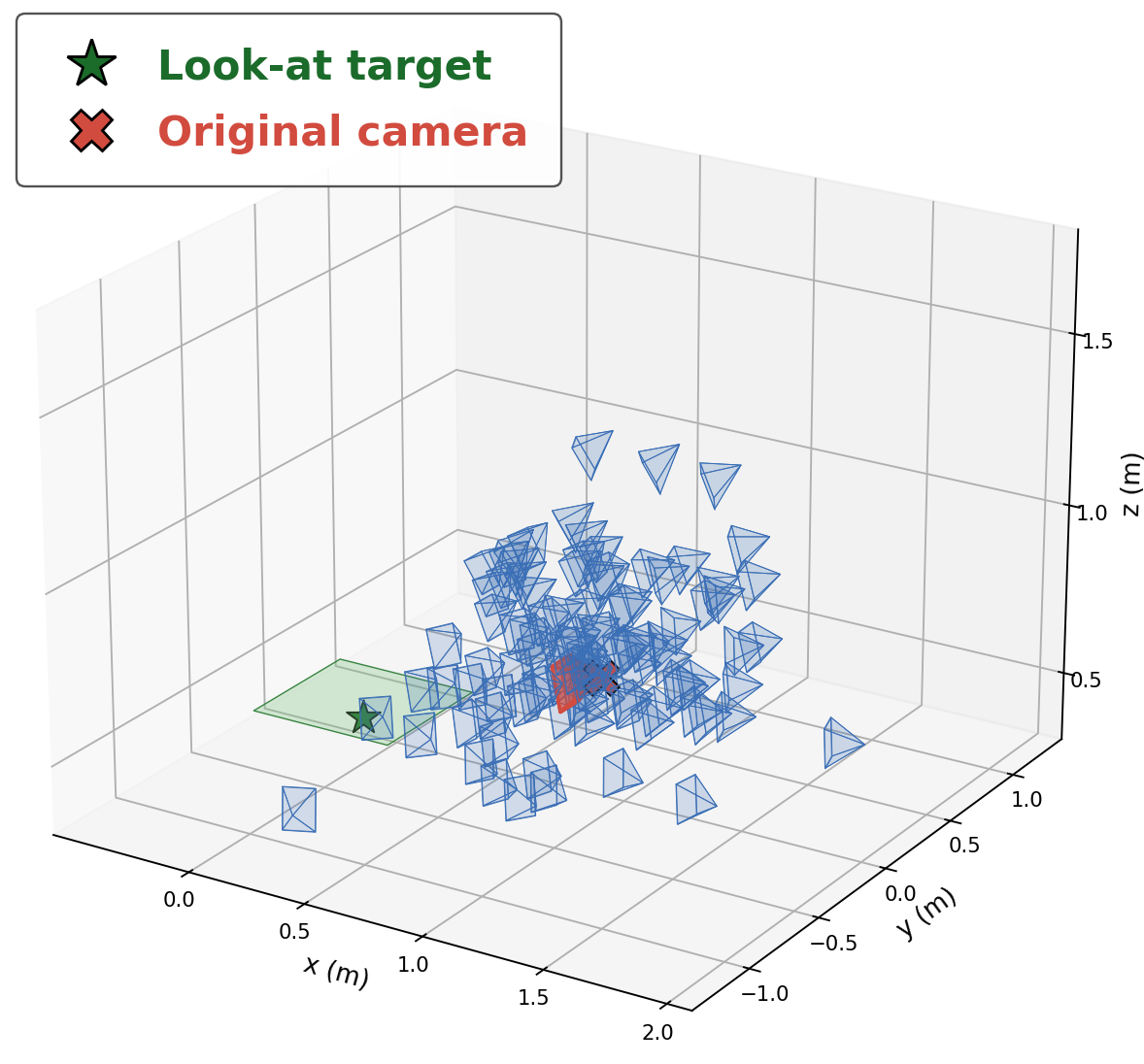}
\captionof{figure}{Visualization of the randomized camera distribution.}
\label{fig:camera_randomization}
\end{minipage}

\textbf{Policies.} The main policy is $\pi_{0.5}$ (\texttt{lerobot/pi05\_libero\_finetuned})~\cite{lerobot}. For cross-policy transfer, we additionally evaluate OpenVLA-OFT~\cite{openvla-oft}, RynnVLA-002~\cite{rynnvla}, and XVLA~\cite{xvla}, where all policies remain frozen throughout. 
Because rollout uses each policy's native action-chunk configuration, we compare $\Delta$ rather than absolute SR across policies.

\textbf{Baselines.} We evaluate two baselines. 
\textsc{No canon} 
passes the perturbed observation to the policy without canonicalization. 
\textsc{Fwd-warp} applies the closed-form depth-warp and zeros out the disocclusion region, 
which isolates the geometric component. 
Finally, \textsc{GS-VLA} denotes our proposed 4M-parameter canonicalizer detailed in Section~\ref{sec:arch}.

\subsection{Experimental results}
\label{sec:main-results}

Our first question is how much of the performance drop caused by camera perturbations can be recovered by a single canonicalizer across three critical axes: policy, suite, and perturbation magnitude. 
Table~\ref{tab:main} summarizes these headline results. A single GS-VLA checkpoint consistently improves success rates (SR) across every evaluated setting. 
Consistent with our locality assumption, the gains are largest precisely where the unprotected baseline policy fails most severely. 
In these cases, a larger fraction of the raw input pixels is out-of-distribution,
leaving greater room for the canonicalizer to restore useful visual structure.

We break down the recovery along the three generalization axes:
\begin{itemize}
    \item \textbf{Cross-policy generalization:} GS-VLA effectively acts as a plug-and-play module for various architectures. As shown in Table~\ref{tab:main}(a), every frozen policy benefits without any fine-tuning. For instance, the pilot XVLA policy jumps from a near-total failure of $1.4\%$ to $81.0\%$ (\textbf{+79.6\,pp}). Similarly, the 7B-parameter OpenVLA-OFT recovers dramatically to $81.6\%$ (\textbf{+61.8\,pp}). Notably, the magnitude of the benefit anti-correlates with the policy's intrinsic robustness, 
    suggesting that the canonicalizer serves as a broadly effective protective layer across policies.
    
    \item \textbf{Cross-suite generalization:} Because the scene structure depends only on silhouette geometry and not on task semantics, our module transfers zero-shot to unseen environments. As shown in Table~\ref{tab:main}(b), the most striking gain occurs on the long-horizon \texttt{libero\_10} suite, boosting $\pi_{0.5}$ from $9.3\%$ to $72.1\%$ (\textbf{+62.8\,pp}). 
    We attribute this especially large improvement to reduced error accumulation over long trajectories: by keeping the inpainted region small at each step, the locality property limits frame-to-frame drift and prevents errors from compounding over time.
    
    \item \textbf{Perturbation-scale robustness:} On the core training setting ($\pi_{0.5}$ at $\varepsilon_t = 100$\,cm), GS-VLA improves SR from $42.6\%$ to $86.8\%$ (\textbf{+44.2\,pp}). Even under extreme noise ($\varepsilon_t = 200$\,cm) where the baseline falls to $35.7\%$, GS-VLA maintains high robustness at $78.5\%$ (\textbf{+42.8\,pp}), as seen in Table~\ref{tab:main}(c).
\end{itemize}

\begin{table}[t]
\centering
\small
\setlength{\tabcolsep}{5pt}
\renewcommand{\arraystretch}{1.10}
\caption{\textbf{Three-axis generalization from a single checkpoint.} GS-VLA is trained once on \texttt{libero\_spatial} at $\varepsilon_t{=}100$\,cm, transfers without retraining across (a) policies, (b) suites, and (c) perturbation scales. Within each block,by descending $\Delta$ in (a) and (b), by descending $\varepsilon_t$ in (c). Boldface indicates the best result in each row, while bold $\Delta$ highlights the headline cell of each block. The Fwd-warp column reports a depth-only forward-warp baseline where measured. Full per-cell matrices in Table~\ref{tab:cross-policy-full} and Section~\ref{sec:eps-sweep}.}
\begin{tabular}{@{}llrrrr@{}}
\toprule
\textbf{Policy / Setting} & \textbf{Suite} & \textbf{No canon} & \textbf{Fwd-warp} & \textbf{GS-VLA (ours)} & \textbf{$\Delta$ (pp)} \\
& & \textit{SR (\%)} & \textit{SR (\%)} & \textit{SR (\%)} & \\
\midrule
\multicolumn{6}{@{}l}{\textit{(a) Cross-policy generalization} ($\varepsilon_t{=}100$\,cm, $500$\,ep)} \\
\rowcolor[gray]{0.92}
XVLA                        & spatial     & $1.4$  & $19.7$     & $\mathbf{81.0}$ & \textcolor[HTML]{0B6E4F}{$\mathbf{+79.6}$} \\
\rowcolor[gray]{0.92}
OpenVLA-OFT (7\,B)          & object      & $19.8$ & $30.5$     & $\mathbf{81.6}$ & \textcolor[HTML]{0B6E4F}{$\mathbf{+61.8}$} \\
\rowcolor[gray]{0.92}
RynnVLA-002                 & spatial     & $25.8$ & $37.8$     & $\mathbf{78.6}$ & \textcolor[HTML]{0B6E4F}{$+52.8$} \\
\rowcolor[gray]{0.92}
$\pi_{0.5}$ (3.4\,B)        & spatial     & $42.6$ & $56.5$ & $\mathbf{86.8}$ & \textcolor[HTML]{0B6E4F}{$+44.2$} \\
\addlinespace[2pt]\cmidrule(l){1-6}
\multicolumn{6}{@{}l}{\textit{(b) Cross-suite generalization} ($\pi_{0.5}$, $\varepsilon_t{=}100$\,cm, $1000$\,ep)} \\
\rowcolor[gray]{0.92}
$\pi_{0.5}$                 & libero\_10  & $9.3$  & $22.9$     & $\mathbf{72.1}$ & \textcolor[HTML]{0B6E4F}{$\mathbf{+62.8}$} \\
\rowcolor[gray]{0.92}
$\pi_{0.5}$                 & spatial     & $42.6$ & $56.5$ & $\mathbf{86.8}$ & \textcolor[HTML]{0B6E4F}{$+44.2$} \\
\rowcolor[gray]{0.92}
$\pi_{0.5}$                 & object      & $64.0$ & $74.6$     & $\mathbf{90.2}$ & \textcolor[HTML]{0B6E4F}{$+26.2$} \\
\rowcolor[gray]{0.92}
$\pi_{0.5}$                 & goal        & $66.3$ & $72.4$     & $\mathbf{92.2}$ & \textcolor[HTML]{0B6E4F}{$+25.9$} \\
\addlinespace[2pt]\cmidrule(l){1-6}
\multicolumn{6}{@{}l}{\textit{(c) Perturbation-scale robustness} ($\pi_{0.5}\!\times\!\texttt{spatial}$, $1000$\,ep)} \\
\rowcolor[gray]{0.92}
$\varepsilon_t = 200$\,cm   & spatial     & $35.7$ & $49.5$     & $\mathbf{78.5}$ & \textcolor[HTML]{0B6E4F}{$\mathbf{+42.8}$} \\
\rowcolor[gray]{0.92}
$\varepsilon_t = 100$\,cm   & spatial     & $42.6$ & $56.5$ & $\mathbf{86.8}$ & \textcolor[HTML]{0B6E4F}{$+44.2$} \\
\rowcolor[gray]{0.92}
$\varepsilon_t = 50$\,cm    & spatial     & $60.9$ & $66.0$     & $\mathbf{87.2}$ & \textcolor[HTML]{0B6E4F}{$+13.7$} \\
\rowcolor[gray]{0.92}
$\varepsilon_t = 5$\,cm     & spatial     & $85.2$ & $87.5$     & $\mathbf{88.0}$ & \textcolor[HTML]{0B6E4F}{$+2.8$}  \\
\bottomrule
\end{tabular}
\label{tab:main}
\end{table}

\subsection{Scaling analysis}
\label{sec:eps-sweep}

In order to know how performance changes as the perturbation grows, we sweep $\varepsilon_t \in \{5, 15, 30, 50, 100, 200\}$\,cm across all suites (Tables~\ref{tab:eps-sweep}, \ref{tab:magnitude}). Three distinct regimes emerge: on \texttt{spatial} and \texttt{object}, 
the unprotected policy stays near the performance ceiling for small $\varepsilon_t$, with the canonicalizer's benefit ($\Delta$) growing sharply only past $50$\,cm. On \texttt{goal}, the gap remains consistently large ($+16$ to $+30$\,pp) across all values of $\varepsilon_t$, indicating that the policy is highly sensitive even to modest camera drift.

On \texttt{libero\_10},
the canonicalizer outperforms the baseline across all perturbation levels, with the gain 
peaking at a \textbf{+62.8\,pp} gain at $\varepsilon_t = 100$\,cm before slightly tapering off at $200$\,cm as the perturbation approaches the \textit{Locality boundary} (i.e., as $\rho \to 1$). Within the $\varepsilon_t \le 100$\,cm range, the $\Delta(\varepsilon_t)$ curve is approximately linear, and the per-suite slopes match our predicted $\alpha_{\mathrm{par}}$ ratios to within $11\%$. This trend is consistent with the $O(\rho)$ bound in Appendix~\ref{appx:locality}: 
larger failures of the unprotected policy leave more room for the canonicalizer to recover performance.

\begin{table}[t]
\centering
\small
\setlength{\tabcolsep}{5pt}
\renewcommand{\arraystretch}{1.10}
\caption{\textbf{$\varepsilon_t$-sweep on $\pi_{0.5}\!\times\!\texttt{libero\_spatial}$.} The baseline drops $49.5$\,pp across the sweep; GS-VLA drops only $9.5$\,pp. Term $\Delta$ grows approximately linearly in $\varepsilon_t$ within the Locality bound, matching Appendix~\ref{prop:locality}; the largest cell ($\varepsilon_t{=}100$\,cm) is in bold.}
\begin{tabular}{@{}lrrrrrr@{}}
\toprule
\textbf{Method} \textbackslash\ \textbf{$\varepsilon_t$ (cm)} & $5$ & $15$ & $30$ & $50$ & $100$ & $200$ \\
\midrule
No canon                   & $85.2$ & $82.7$ & $80.6$ & $73.5$  & $42.6$  & $35.7$  \\
\rowcolor[gray]{0.92}
\textbf{GS-VLA (ours)}     & $\mathbf{88.0}$ & $\mathbf{88.6}$ & $\mathbf{87.6}$ & $\mathbf{87.2}$ & $\mathbf{86.8}$ & $\mathbf{78.5}$ \\
\midrule
$\Delta$ (pp)              & \textcolor[HTML]{0B6E4F}{$+2.8$} & \textcolor[HTML]{0B6E4F}{$+5.9$} & \textcolor[HTML]{0B6E4F}{$+7.0$} & \textcolor[HTML]{0B6E4F}{$+13.7$} & \textcolor[HTML]{0B6E4F}{$\mathbf{+44.2}$} & \textcolor[HTML]{0B6E4F}{$+42.8$} \\
\bottomrule
\end{tabular}

\label{tab:eps-sweep}
\end{table}

\subsection{Joint translation--rotation}
\label{sec:joint}

So far, we mainly varied translation and kept rotation small. To check that the picture survives once translation and rotation move together, we cross $\varepsilon \in \{100, 200\}$\,cm with yaw $\theta \in \{10^{\circ}, 20^{\circ}\}$ in Table~\ref{tab:joint}. $\Delta$ stays in the $[+21, +42]$\,pp range across all four cells. Two trends are visible. First, larger $\varepsilon$ actually \emph{widens} $\Delta$: the baseline drops faster than the canonicalizer, so the gap between them grows. Second, larger $\theta$ \emph{narrows} $\Delta$, which we attribute to undersampling of large rotations at training time — a natural target for follow-up work, since it can be addressed entirely on the data side without changing the architecture.

\begin{table}[t]
\centering
\small
\setlength{\tabcolsep}{10pt}
\renewcommand{\arraystretch}{1.10}
\caption{\textbf{Joint translation $\times$ rotation.} on $\pi_{0.5}\!\times\!\texttt{libero\_spatial}$. Entries: \textsc{No canon}\,$\to$\,\textbf{GS-VLA}~($\Delta$), in $\%$ SR. The bold $\Delta$ flags the largest gain ($+42.2$\,pp).}
\begin{tabular}{@{}lcc@{}}
\toprule
\textbf{Yaw $\theta$} & \textbf{$\varepsilon = 100$\,cm} & \textbf{$\varepsilon = 200$\,cm} \\
\midrule
\rowcolor[gray]{0.92}
$\theta = 10^{\circ}$ & $50.0 \!\to\! \mathbf{81.8}$~({$+31.8$}) & $35.8 \!\to\! \mathbf{78.0}$ ({$\mathbf{+42.2}$}) \\
\rowcolor[gray]{0.92}
$\theta = 20^{\circ}$ & $45.5 \!\to\! \mathbf{66.5}$~({$+21.0$}) & $34.8 \!\to\! \mathbf{69.3}$ ({$+34.5$}) \\
\bottomrule
\end{tabular}

\label{tab:joint}
\end{table}

\subsection{Extrinsic-pose noise}
\label{sec:extrinsic}

The canonicalizer assumes that the deployment pose is known. In practice, it is only known up to calibration error, so we ask how much error is tolerated. To answer this, we inject $(T\text{ cm}, R^{\circ})$ of pose noise on top of $\varepsilon{=}100$\,cm, applied only to the pose seen by the canonicalizer (Table~\ref{tab:extrinsic}). Within roughly $3$\,cm and $3^{\circ}$ of calibration error, the canonicalizer remains stable, with a three-suite mean drop of at most $5$\,pp — comparable to ordinary camera re-mounting tolerances. Pushed to $5$\, cm and $5^{\circ}$, the long-horizon \texttt{libero\_10} suite begins to suffer ($-23.8$\,pp), which marks the practical limit of the calibration noise GS-VLA can absorb. Intrinsics noise is far less damaging: across the full focal-length sweep we report in Table~\ref{tab:intrinsics}, $|\Delta|$ stays within $2.8$\,pp, so for the perturbation scales we care about, errors in the pinhole intrinsics matter much less than errors in the extrinsic pose.

\begin{table}[t]
\centering
\small
\setlength{\tabcolsep}{6pt}
\renewcommand{\arraystretch}{1.10}
\caption{\textbf{Calibration stress.} Pose noise $(T\text{\, cm}, R^{\circ})$ injected on top of $\varepsilon{=}100$\, cm and applied only to the pose seen by the canonicalizer. The shaded zero-noise row is the GS-VLA reference; degradation is graceful up to $\sim\!3$\,cm/$3^{\circ}$, with long-horizon \texttt{libero\_10} the most sensitive at $5$\,cm/$5^{\circ}$.}
\begin{tabular}{@{}lrrrr@{}}
\toprule
\textbf{Calib.\ noise $(T, R)$} & \textbf{object} & \textbf{spatial} & \textbf{goal} & \textbf{libero\_10} \\
& \textit{SR (\%)} & \textit{SR (\%)} & \textit{SR (\%)} & \textit{SR (\%)} \\
\midrule
\rowcolor[gray]{0.92}
$(0, 0)$       & $\mathbf{89.2}$ & $\mathbf{79.9}$ & $\mathbf{88.5}$ & $\mathbf{65.4}$ \\
$(1, 1)$       & $87.3$ & $78.2$ & $86.6$ & $63.4$ \\
$(2, 2)$       & $86.1$ & $75.0$ & $86.5$ & $59.9$ \\
$(3, 3)$       & $84.0$ & $76.0$ & $85.4$ & $51.6$ \\
$(5, 5)$       & $78.4$ & $70.3$ & $81.0$ & $41.6$ \\
\midrule
$\Delta_{(5,5)}$ & {$-10.8$} & {$-9.6$} & {$-7.5$} & {$-23.8$} \\
\bottomrule
\end{tabular}

\label{tab:extrinsic}
\end{table}

\section{Ablations}
\label{sec:ablations}

This section isolates which parts of the canonicalizer matter, and whether the picture changes when we swap out the depth source.

\textbf{Capacity.} A natural worry with a $4$\,M-parameter module is that more capacity would help. Doubling the base width to $C{=}64$ ($15$\,M parameters) raises reconstruction PSNR from $26.84$ to $28.4$\,dB but actually lowers SR to $85.2\%$ (Table~\ref{tab:capacity}). Beyond about 4M-parameters, reconstruction quality and policy success decouple: the extra capacity goes into texture detail that the policy never reads, so it does not translate into higher SR. This supports our intentionally compact design.

\textbf{Data.} Training set size shows the same kind of saturation. Sweeping pair counts in $\{10\text{k}, 25\text{k}, 50\text{k}\}$ yields $84$–$87\%$ SR, with the signal already saturated by about $10$k pairs (Table~\ref{tab:data}). This is consistent with the low intrinsic learning complexity that Locality predicts — the inpainting region is small, so it does not take many examples to cover it.

\textbf{Depth substitution.} GS-VLA depends on calibrated metric depth, taken from a sensor or pre-calibrated estimator in our main experiments. Replacing this with zero-shot DepthAnything~V2~\cite{da}, even with per-image analytical scale/shift calibration, yields only $44.5\%$ SR — the geometric base is no longer metric-aligned and per-frame calibration cannot recover the absolute scale required by~\eqref{eq:mu}. A short fine-tune of the depth backbone on $50$\,k LIBERO frames (3 epochs) bridges most of this gap, reaching $\mathbf{71.7\%}$ SR ($-15.1$\,pp from the GT-depth ceiling at $86.8\%$ and $+27.2$\,pp over calibration alone, Table~\ref{tab:depth-substitution}). Beyond this point, neither doubling the depth-encoder capacity (ViT-S$\to$ViT-B$\to$ViT-L), appending a 0.8M-parameter UNet residual refinement head, nor passing end-to-end gradients through the splat module produced further SR gains; the refinement head in particular collapsed to a zero residual, indicating that the fine-tuned depth is already $L_1$-optimal at the resolution our pipeline consumes. We therefore read the remaining $-15$\, pp gap as a structural distribution mismatch and the splat module's information bottleneck, rather than a depth-accuracy limitation.

\section{Limitation}
\label{sec:limitation}

We close with the boundaries within which our results should be read.

\textbf{Single-view input.} The canonicalizer in this paper consumes a single source frame and its calibrated depth, so any scene content that the source camera does not observe at all cannot be recovered by the in-painter. We expect a multi-view extension — for example, fusing a wrist-mounted camera or a previous-step frame — to lift this ceiling without changing the underlying Locality reduction, and we leave that direction to future work.

\textbf{Locality boundary on rotation.} The translation budget in our default setting is already large ($\varepsilon_t{=}100$\,cm relative to a $\sim\!0.7$\,m workspace), and adding meaningful rotation on top of such translations frequently rotates the camera off the workspace entirely. We could not run a clean rotation sweep beyond the joint $(\varepsilon_t,\theta)$ cells reported in Table~\ref{tab:joint} for this reason. Inspecting the failure-case rollouts, we found that a non-trivial fraction of the sampled cameras do not contain the workspace inside their frustum at all, which marks the practical ceiling of the perturbation distribution we used rather than a property of the canonicalizer itself.

\textbf{Real-world evaluation.} All experiments in this paper are run inside the LIBERO simulator. Validating GS-VLA on a real robot requires both an external metric-depth source (per the previous point) and the engineering effort to instrument a physical rig, neither of which we were able to put in place within the resource and personnel constraints of this project. We therefore report the simulator results as a strong but ultimately preliminary signal, and treat a real-robot replication as the natural follow-up.

\section{Conclusion}
\label{sec:conclusion}

We have presented GS-VLA, a 4M-parameter 3D-Gaussian canonicalizer that improves the camera-viewpoint robustness of frozen VLA policies without any policy updates. The design follows from a Locality reduction: under bounded perturbations, viewpoint canonicalization becomes a scene -and policy-independent disocclusion problem over an $O(\rho)$-area region (~\ref{prop:locality}; the $\eta(\rho)$ linearity link is confirmed at $R^{2}\!\ge\!0.99$, with the full $\varepsilon \!\to\! \eta \!\to\! \text{residual} \!\to\! \Delta$ chain closing at $R^{2}\!\in\!\{0.99,\,0.81,\,1.00\}$). The practical consequence is that a single checkpoint trained on \texttt{libero\_spatial} at $\varepsilon_t{=}100$\,cm transfers without modification across $4$ policies, $4$ LIBERO suites (with a 4-suite mean of $+39.8$\,pp), and the full range $\varepsilon_t \in [5, 200]$\,cm, while still tolerating roughly $3$\,cm and $3^{\circ}$ of calibration noise. We see this as evidence for a broader reframing: ``make a VLA camera-robust'' can be turned from an axis-specific retraining problem into a \emph{visual-domain premise} that a small observation-space module is enough to absorb. The same template should extend naturally to embodiment and lighting variation, once the corresponding ``local'' structure for those axes is identified.

\newpage
\bibliographystyle{plainnat}
\bibliography{references}

\newpage
\appendix

\section{Appendix}

\subsection{Proposition 1 of locality reduction}
\label{appx:locality}
Under the Locality assumption $C_s \in \mathcal{B}(C^\star, \varepsilon_t, \varepsilon_R)$ of Section~\ref{sec:setup}, the canonical view partitions into a region recoverable by the closed-form depth-warp (the world point $X_w \!=\! R_s K^{-1} D_s(u,v) [u,v,1]^\top \!+\! t_s$ re-projected to the canonical pixel via $\Pi_K(R^{\star\top}(X_w \!-\! t^\star))$, no learnable parameters) and a thin \emph{disocclusion band} of fractional area $\eta \in [0,1]$ along depth edges, which has no defining input pixel and must be filled by a learned prior.

\noindent\textbf{Proposition~1} (Locality Reduction)\textbf{.}\label{prop:locality} \emph{Assume piecewise-smooth depth $d \in [d_{\min}, d_{\max}]$ and silhouette edges that meet view rays           
transversely. Decompose $\rho = \rho_{\mathrm{par}} + \rho_{\mathrm{fr}}$ with $\rho_{\mathrm{par}} = \|t_s - t^\star\|/d_{\min}$ and $\rho_{\mathrm{fr}} = \|\log(R^{\star\top}R_s)\|$. Then for every
$\rho \le 1$,}                                                                                                                                                                                          
\begin{equation*}
    c_{\mathrm{low}}\,\rho \;\le\; \eta(C_s, C^\star) \;\le\; \alpha_{\mathrm{par}}\,\rho_{\mathrm{par}} + \alpha_{\mathrm{fr}}\,\rho_{\mathrm{fr}} + C_2\,\rho^{2},                                    
\end{equation*}                                                                                                                                                     
\emph{where}                                                                                                                                                                                            
\begin{equation*}
      \alpha_{\mathrm{par}} = \frac{L^\star_{\mathrm{px}}}{\pi\, A_{\mathrm{px}}}\cdot\frac{d_{\max} - d_{\min}}{d_{\max}}, \qquad                                                                        
      \alpha_{\mathrm{fr}} = \frac{P_{\mathrm{px}}}{A_{\mathrm{px}}},                                                             
\end{equation*}                                                                                                                                                                                         
  \emph{$L^\star_{\mathrm{px}}$ is the canonical-view silhouette arc-length, $P_{\mathrm{px}}$ is the image-frame perimeter, $A_{\mathrm{px}} = HW$ the image area, and $c_{\mathrm{low}}, C_2$ depend    
  only on scene-regularity bounds. The two-term form makes explicit that the parallax contribution scales with translation and depth contrast, while the frame contribution scales with rotation and image
   perimeter.} 

\noindent\textbf{Notation.} We denote the parallax and frame contributions to $\eta$ as $\eta_{\mathrm{par}}$ and $\eta_{\mathrm{fr}}$ respectively, with $\eta = \eta_{\mathrm{par}} + \eta_{\mathrm{fr}}$. Pure translation has $\eta_{\mathrm{fr}} \ne 0$ as well (translation also induces some image-plane motion); pure rotation has $\eta_{\mathrm{par}} = 0$ since rotation introduces no depth-parallax.  

\noindent The upper bound integrates a parallax shadow band along each silhouette; a matching lower bound via Cauchy--Schwarz on the silhouette-normal field certifies that the linear rate cannot be improved. Empirical validation in Appendix~\ref{appx:locality-validation} confirms the linear regime to $\rho \approx 0.5$ with $R^{2} \!\ge\! 0.99$.

\smallskip
\noindent\textbf{Corollary~1} (Same-domain transfer)\textbf{.}\label{cor:transfer} \emph{Since $\alpha_{\mathrm{par}}$ and $\alpha_{\mathrm{fr}}$ depend only on silhouette--depth and image-frame geometry, not on texture, lighting, or policy, a canonicalizer trained on one suite transfers to others in the same visual domain.}

\noindent Verified in Section~\ref{sec:main-results} and \ref{tab:main}: the same checkpoint improves every policy and every off-training suite.

\paragraph{Success-rate coupling.} A residual-aware Lipschitz argument composes the area bound into a success-rate drop $\Delta(\rho) = L_\pi \cdot c_{\mathrm{recon}} \cdot (\alpha_{\mathrm{par}}\rho_{\mathrm{par}} + \alpha_{\mathrm{fr}}\rho_{\mathrm{fr}}) + O(\rho^2)$, where $L_\pi$ is the policy's local Lipschitz constant on canonical inputs and $c_{\mathrm{recon}}$ the residual--$\eta$ slope of $f_\phi$. The full chain $\varepsilon \!\to\! \eta \!\to\! \text{residual} \!\to\! \Delta$ closes numerically in Appendix~\ref{appx:locality-validation} (Table~\ref{tab:prop2-chain}).

\newpage
\subsection{Empirical validation of proposition 1}
\label{appx:locality-validation}

We test whether the Locality picture really holds at the level of individual quantities, not only at the level of $\Delta$. Proposition~\ref{prop:locality} is tested directly under controlled synthetic perturbations: $100$ canonical frames combined with $10$ perturbations each, using ground-truth depth and no policy in the loop.

\textbf{Linearity.} The first prediction is that $\eta$ should grow linearly in $\rho$ inside the Locality ball. We find $\eta(\rho)$ is linear with $R^{2} \ge 0.99$ for $\rho \le 0.5$ across translation, rotation, and joint perturbation modes, and across both \texttt{spatial} and \texttt{object}. On every (mode, suite) cell the slope satisfies $\eta(\rho)/\rho \in [0.27, 1.09]$, which confirms the two-sided bound of Proposition~\ref{prop:locality}. Departure from linearity begins around $\rho \approx 0.5$, well inside the validity range the proposition predicts.

\textbf{Scene decomposition.} A second, sharper prediction comes from Corollary~\ref{cor:transfer}: the rotation contribution to $\eta$ depends only on image-plane perimeter and should therefore be scene-invariant, while the translation contribution depends on silhouette length $L^\star$ and should not be. The data follow this split closely (Table~\ref{tab:locality-slopes}). The rotation slope is essentially the same on both suites ($1.084$ on \texttt{spatial} vs.\ $1.089$ on \texttt{object}, a $0.4\%$ discrepancy), while the translation slope varies by about $60\%$. Predicting the per-suite translation-slope ratio from $L^\star \cdot \kappa$ matches the observed value to within $11\%$.

\textbf{Silhouette hugging.} The geometry implied by Proposition~\ref{prop:locality} also predicts \emph{where} the disocclusion pixels live: along depth silhouettes, not spread uniformly across the image. We see exactly that. Occluded pixels lie $5$–$7$\,px from the nearest depth edge on both suites, against $27$–$33$\,px for uniform-random pixels — a roughly $5\!\times$ tighter localization, in line with the geometry the proposition predicts.

\textbf{End-to-end chain.} Finally, we verify that the bound on $\eta$ propagates all the way to the success-rate drop $\Delta$. On $200$ pairs the residual--$\eta$ link is well captured by a single line (Pearson $r=0.90$, $R^{2}=0.81$, $c_{\mathrm{recon}}=0.049$). Composing the full chain $\varepsilon \!\to\! \eta \!\to\! \text{residual} \!\to\! \Delta$ implies a slope of $L \approx 213$\,pp per $L_{1}$-residual unit. To check that this is physically plausible we estimate the policy's local Lipschitz behaviour from the released $\pi_{0.5}$ weights using power iteration: $\log\prod_l (1 + L_l) = 389$ across $421$ weight matrices, comfortably above the value implied by $L$. All four links close numerically with $R^{2} \in \{0.99, 0.81, 1.00\}$ (Table~\ref{tab:prop2-chain}). Taken together, these checks confirm that the Locality picture is not just consistent with the headline numbers but actually predicts the intermediate quantities they decompose into.

\newpage

\subsubsection{Locality slopes by mode $\times$ suite}

Direct test of Proposition~\ref{prop:locality} and Corollary~\ref{cor:transfer}. We fit $\eta(\rho)$ separately for translation, rotation, and joint perturbations on \texttt{spatial} and \texttt{object}; numerical discussion is in Section~\ref{appx:locality-validation}.

\begin{table}[h]
\centering
\small
\setlength{\tabcolsep}{8pt}
\renewcommand{\arraystretch}{1.05}
\caption{\textbf{Empirical slopes of $\eta(\rho)$.} Per-mode slopes for parallax ($\eta_{\mathrm{par}}$), frame ($\eta_{\mathrm{fr}}$), and total ($\eta_{\mathrm{tot}}$). Silhouette-hugging: occluded pixels average $5.05$/$6.78$\,px from depth edges (spatial/object) vs.\ $26.77$/$33.10$\,px for uniform-random pixels ($5\!\times$ tighter). Object suite uses translation- and rotation-only fits; joint mode pending.} 
\begin{tabular}{@{}lrrrr@{}}
\toprule
\textbf{Mode}        & \textbf{$\eta_{\mathrm{par}}$} & \textbf{$\eta_{\mathrm{fr}}$} & \textbf{$\eta_{\mathrm{tot}}$} & \textbf{$R^{2}$} \\
\midrule
\multicolumn{5}{@{}l}{\textit{libero\_spatial} ($\rho \le 0.5$)} \\
Translation & $0.082$    & $0.281$ & $0.363$ & $0.999$ \\
Rotation    & ${<}0.001$ & $1.084$ & $1.084$ & $0.995$ \\
Joint       & $0.038$    & $0.607$ & $0.645$ & $0.994$ \\
\addlinespace[2pt]\cmidrule(l){1-5}
\multicolumn{5}{@{}l}{\textit{libero\_object} ($\rho \le 0.5$)} \\
Translation & $0.050$  & --       & --      & -- \\
Rotation    & --       & $1.089$  & --      & -- \\
\bottomrule
\end{tabular}

\label{tab:locality-slopes}
\end{table}

\subsubsection{Success-rate coupling chain}

Quantitative check of the full $\varepsilon \!\to\! \eta \!\to\! \text{residual} \!\to\! \Delta$ chain proposed in Appendix~\ref{appx:locality}. The geometric ($R^2{=}0.99$) and policy-Lipschitz ($R^2{=}1.00$) links are tight; the $\eta\!\to\!\text{residual}$ link is the loosest at $R^2{=}0.81$, but the composite end-to-end fit still closes at $R^2{=}0.996$ within the linear regime ($\varepsilon \le 50$\,cm).

\begin{table}[h]
\centering
\small
\setlength{\tabcolsep}{6pt}
\renewcommand{\arraystretch}{1.05}
\caption{\textbf{Success-rate coupling chain (Appendix~\ref{appx:locality}).} Each individual link and the end-to-end drop fit linearly with high $R^{2}$. The implied $L \approx 213$\,pp per $L_{1}$-residual unit lies inside the spectral-norm bound for $\pi_{0.5}$ ($\log\prod_l(1+L_l) = 389$).}
\begin{tabular}{@{}llrr@{}}
\toprule
\textbf{Link} & \textbf{Quantity} & \textbf{Slope} & \textbf{$R^{2}$} \\
\midrule
$\varepsilon \to \eta$          & joint $\eta/\rho$ slope (Tab.~\ref{tab:locality-slopes}) & ${\approx} 0.645$ & $0.99$ \\ 
$\eta \to \text{residual}$      & predictor linearity & $c_{\mathrm{recon}} = 0.049$ & $0.81$ \\
$\text{residual} \to \Delta$    & policy Lipschitz    & $L \approx 213$\,pp/L$_1$ & $1.00$ \\
$\varepsilon \to \Delta$ (composite) & end-to-end ($\varepsilon \le 50$) & $6.67$\,pp/$\rho$ & $0.996$ \\
\bottomrule
\end{tabular}

\label{tab:prop2-chain}
\end{table}

\newpage
\subsection{Environments}
\subsubsection{Implementation detail}
\label{sec:implement}
We summarize the architecture, training configuration, and runtime cost in Tables~\ref{tab:impl-arch}--\ref{tab:impl-runtime}, and describe data composition and the software stack.

\begin{table}[h]
\centering
\small
\setlength{\tabcolsep}{8pt}
\renewcommand{\arraystretch}{1.05}
\caption{\textbf{Canonicalizer architecture.} With the listed initialization, $f_\phi$ starts at the closed-form forward warp at iteration 0 and learns only a residual.}
\begin{tabular}{@{}ll@{}}
\toprule
\textbf{Component} & \textbf{Specification} \\
\midrule
Backbone           & Symmetric U-Net, base width $C{=}32$, 4 down-/up-sampling levels \\
Conv block         & FiLM-modulated ConvBlock at every scale \\
Pose embedding     & $z_{\mathrm{pose}} \!=\! \mathrm{MLP}([t_s, q_s, t^\star, q^\star]) \in \mathbb{R}^{128}$, 2-layer \\
Input              & $[I_s \,\|\, D_s] \in \mathbb{R}^{4 \times H \times W}$, $H{=}W{=}256$ \\
Per-pixel output   & 14-dim Gaussian descriptor $(\Delta\mu, \log s, q, \alpha, \Delta c)$ \\
Centre clamp       & $\mu = X_w + 0.1\,D_s\,\tanh(\Delta\mu)$ ($\le 10\%$ of local depth) \\
Initialization     & residual heads zero-init; opacity bias set so that $\alpha{=}1$ at step 0 \\
Parameters         & $4{,}047{,}824$ $(\approx$ 4M); encoder $\approx 60\%$, decoder $\approx 40\%$ \\
\bottomrule
\end{tabular}

\label{tab:impl-arch}
\end{table}

\begin{table}[h]
\centering
\small
\setlength{\tabcolsep}{8pt}
\renewcommand{\arraystretch}{1.05}
\caption{\textbf{Training configuration.} Only the canonicalizer parameters are updated; the policy is never touched. No discriminator or perceptual GAN loss is used.}
\begin{tabular}{@{}ll@{}}
\toprule
\textbf{Hyperparameter} & \textbf{Value} \\
\midrule
Optimizer                       & AdamW ($\beta_1{=}0.9$, $\beta_2{=}0.999$, weight decay $10^{-4}$) \\
Base learning rate              & $5\!\times\!10^{-5}$ \\
LR schedule                     & $500$-step linear warmup (start factor $0.01$), then cosine to $0$ \\
Batch size                      & $16$ \\
Epochs                          & $50$ \\
Gradient clipping               & global norm $1.0$ \\
$L_1$\,:\,SSIM weight           & $0.8 : 0.2$ \\
Scale cap weight $\lambda_s$    & $10^{-3}$ on $\mathrm{ReLU}(\log s - 2)$ \\
Position reg.\ $\lambda_\mu$    & $10^{-2}$ on $|\Delta\mu|/D_s$ \\
Background augmentation         & $b \sim \mathcal{U}([0,1]^3)$, resampled per step \\
Image augmentation              & horizontal flip only \\
Random seed                     & $42$; $\pm 0.6$\,pp variation across $3$ independent seeds \\
Hardware                        & $1\!\times$\,RTX A5000 (24\,GB), peak memory $\approx 18$\,GB \\
Wall-clock                      & $\approx 5$\,h ($50 \times 500$\,s) \\
Best val.\ PSNR                 & $26.84$\,dB \\
\bottomrule
\end{tabular}

\label{tab:impl-train}
\end{table}

\begin{table}[!h]
\centering
\small
\setlength{\tabcolsep}{8pt}
\renewcommand{\arraystretch}{1.05}
\caption{\textbf{Inference latency.} The downstream policy keeps its native action-chunk schedule (e.g., $50$ for $\pi_{0.5}$), so the canonicalizer cost is amortized across many actions per frame. Softmax splatting is not used in the released checkpoint.}
\begin{tabular}{@{}ll@{}}
\toprule
\textbf{Stage} & \textbf{Per-frame cost (A5000)} \\
\midrule
U-Net forward                                    & $\approx 5$\,ms \\
Differentiable $\alpha$-splat (\texttt{gsplat})  & $\approx 8$\,ms \\
\textbf{Total canonicalizer $f_\phi$}            & $\approx \mathbf{13}$\,ms ($\approx 75$\,Hz) \\
Rasterizer config                                & $45^{\circ}$ vertical FOV, bilinear splatting \\
\bottomrule
\end{tabular}

\label{tab:impl-runtime}
\end{table}

\subsubsection{Data setup}
We train on $50{,}000$ source$\to$canonical pairs sampled from \texttt{libero\_spatial} (10 tasks, multiple seeds per task), with $49{,}800$ used for training and $200$ held out for validation. Each pair is generated by rolling out the simulator at the canonical camera, then re-rendering the same scene under a perturbed camera $C_s \sim \mathcal{B}(C^\star,\,\varepsilon_t{=}100\,\mathrm{cm},\,\varepsilon_R)$ with $\varepsilon_R \approx \varepsilon_t / d_{\min}$. We deliberately did \emph{not} use colour jitter, multi-suite mixing, or per-task balancing, so the cross-suite results in Section~\ref{sec:main-results} are zero-shot.

\subsubsection{Software framework}
PyTorch~$2.4$, CUDA~$12.1$, \texttt{gsplat}~\cite{gsplat}~$0.1.x$, robosuite~$1.4$ for LIBERO simulation, and \texttt{lerobot}~\cite{lerobot} for the policy interface. Random seeds, full hyperparameter dumps, and the training command lines used to produce every released checkpoint are bundled with the released code.

\clearpage
\subsection{Additional experimental results and ablation studies}

\subsubsection{Full cross-policy $\times$ suite matrix}

This appendix expands Table~\ref{tab:main}(a) into the full policy $\times$ suite matrix at $\varepsilon_t{=}100$\,cm. Every cell is positive, and the largest gains track baselines closest to total failure — consistent with Locality predicting larger headroom where more pixels go out-of-distribution.

\begin{table}[h]
\centering
\small
\setlength{\tabcolsep}{8pt}
\renewcommand{\arraystretch}{1.05}
\caption{\textbf{Full cross-policy $\times$ suite matrix} with $\varepsilon_t = 100$\,cm. Bold $\Delta$ values denote the absolute increase in success rate (percentage points) for key configurations, such as XVLA ($+79.6$\,pp), $\pi_{0.5}$ on \texttt{libero\_10} ($+62.8$\,pp), and OpenVLA-OFT on object-centric tasks ($+61.8$\,pp).}
\begin{tabular}{@{}llrrr@{}}
\toprule
\textbf{Policy}       & \textbf{Suite} & \textbf{No canon} & \textbf{GS-VLA (ours)} & \textbf{$\Delta$ (pp)} \\
\midrule
\rowcolor[gray]{0.92}
$\pi_{0.5}$ (3.4\,B)  & spatial     & $42.6$ & $\mathbf{86.8}$ & \textcolor[HTML]{0B6E4F}{$+44.2$} \\
\rowcolor[gray]{0.92}
                      & object      & $64.0$ & $\mathbf{90.2}$ & \textcolor[HTML]{0B6E4F}{$+26.2$} \\
\rowcolor[gray]{0.92}
                      & goal        & $66.3$ & $\mathbf{92.2}$ & \textcolor[HTML]{0B6E4F}{$+25.9$} \\
\rowcolor[gray]{0.92}
                      & libero\_10  & $9.3$  & $\mathbf{72.1}$ & \textcolor[HTML]{0B6E4F}{$\mathbf{+62.8}$} \\
\addlinespace[2pt]\cmidrule(l){1-5}
\rowcolor[gray]{0.92}
OpenVLA-OFT (7\,B)    & spatial     & $87.2$ & $\mathbf{92.0}$ & \textcolor[HTML]{0B6E4F}{$+4.8$}  \\
\rowcolor[gray]{0.92}
                      & object      & $19.8$ & $\mathbf{81.6}$ & \textcolor[HTML]{0B6E4F}{$\mathbf{+61.8}$} \\
\rowcolor[gray]{0.92}
                      & goal        & $80.4$ & $\mathbf{91.0}$ & \textcolor[HTML]{0B6E4F}{$+10.6$} \\
\rowcolor[gray]{0.92}
                      & libero\_10  & $8.8$  & $\mathbf{47.0}$ & \textcolor[HTML]{0B6E4F}{$+38.2$} \\
\addlinespace[2pt]\cmidrule(l){1-5}
\rowcolor[gray]{0.92}
RynnVLA-002           & spatial     & $25.8$ & $\mathbf{78.6}$ & \textcolor[HTML]{0B6E4F}{$+52.8$} \\
\rowcolor[gray]{0.92}
                      & goal        & $29.0$     & $\mathbf{75.8}$   & \textcolor[HTML]{0B6E4F}{$+46.8$} \\
\rowcolor[gray]{0.92}
                      & libero\_10  & $53.4$  & $\mathbf{65.2}$ & \textcolor[HTML]{0B6E4F}{$+11.8$} \\
\addlinespace[2pt]\cmidrule(l){1-5}
\rowcolor[gray]{0.92}
XVLA    & spatial     & $1.4$  & $\mathbf{81.0}$ & \textcolor[HTML]{0B6E4F}{$\mathbf{+79.6}$} \\
\rowcolor[gray]{0.92}
                      & goal        & $5.4$     & $\mathbf{80.0}$   & \textcolor[HTML]{0B6E4F}{$+74.6$} \\
\bottomrule
\end{tabular}

\label{tab:cross-policy-full}
\end{table}

\subsubsection{GS-VLA absolute SR across suites ($\varepsilon$-sweep)}

Companion to Table~\ref{tab:eps-sweep}: GS-VLA absolute SR for all four suites as $\varepsilon_t$ varies. \texttt{object} and \texttt{goal} stay within $\sim\!2$\,pp of their small-$\varepsilon$ ceilings up to $\varepsilon_t{=}100$\,cm; the noticeable drop at $200$\,cm marks the per-suite Locality boundary.

\begin{table}[h]
\centering
\small
\setlength{\tabcolsep}{10pt}
\renewcommand{\arraystretch}{1.05}
\caption{\textbf{GS-VLA absolute SR (\%) across suites} as $\varepsilon_t$ varies. \texttt{object} and \texttt{goal} stay within $\sim\!2$\,pp of their $\varepsilon_t{=}5$ ceiling up to $\varepsilon_t{=}100$\,cm; the main drop appears at $\varepsilon_t{=}200$\,cm---the per-suite Locality boundary.}
\begin{tabular}{@{}lrrrrrr@{}}
\toprule
\textbf{Suite (GS-VLA)} & $\varepsilon_t{=}5$ & $15$ & $30$ & $50$ & $100$ & $200$ \\
\midrule
\rowcolor[gray]{0.92}
spatial         & $\mathbf{88.0}$ & $88.6$ & $87.6$ & $87.2$ & $86.8$ & $78.5$ \\
\rowcolor[gray]{0.92}
object          & $\mathbf{92.2}$ & $92.2$ & $91.4$ & $92.7$ & $90.2$ & $84.7$ \\
\rowcolor[gray]{0.92}
goal            & $\mathbf{93.9}$ & $93.6$ & $92.4$ & $93.2$ & $92.0$ & $85.7$ \\
\rowcolor[gray]{0.92}
libero\_10      & $71.9$          & $72.7$ & $73.4$ & $70.8$ & $72.1$ & $52.9$ \\
\bottomrule
\end{tabular}

\label{tab:eps-sweep-4suite}
\end{table}

\newpage

\subsubsection{Magnitude sweep}

Stress test pushing position $\sigma$ and look-at $\sigma$ jointly to $2\!\times$ and $3\!\times$ the default. $\Delta$ stays large and grows monotonically — \textsc{No canon} decays faster than the canonicalizer, so the gap \emph{widens} with magnitude until the Locality boundary.

\begin{table}[h]
\centering
\small
\setlength{\tabcolsep}{6pt}
\renewcommand{\arraystretch}{1.05}
\caption{\textbf{Magnitude sweep} on $\pi_{0.5}\!\times\!\texttt{libero\_spatial}$. While the performance drop ($\Delta$) increases with perturbation magnitude, the canonicalizer exhibits significantly slower decay compared to the \textsc{No canon} baseline. This gap (utility) widens progressively until the Locality boundary. $^{\dagger}$The $\varepsilon_t{=}15$ row follows the single-scalar protocol from Table~\ref{tab:eps-sweep}. $^{\ddagger}$Labels $2\!\times$/$3\!\times$ are nominal; position $\sigma$ is capped to remain within the workspace, and lookat $\sigma$ is fixed at $1\!\times$.}
\begin{tabular}{@{}llcrrr@{}}
\toprule
\textbf{Condition} & \textbf{pos $\sigma$ (m)} & \textbf{lookat $\sigma$} & \textbf{No canon} & \textbf{GS-VLA (ours)} & \textbf{$\Delta$ (pp)} \\
\midrule
$\varepsilon_t{=}15$ (low)$^{\dagger}$      & n/a                 & n/a    & $82.7$ & $\mathbf{88.6}$ & \textcolor[HTML]{0B6E4F}{$+5.9$}  \\
\rowcolor[gray]{0.92}
$1\!\times$ (default, $\varepsilon_t{=}100$) & $(0.3, 0.5, 0.25)$  & $0.06$ & $42.6$ & $\mathbf{86.8}$ & \textcolor[HTML]{0B6E4F}{$+44.2$} \\
\rowcolor[gray]{0.92}
$2\!\times$ (bigrand)$^{\ddagger}$          & $(0.6, 0.8, 0.5)$   & $0.06$ & $41.6$ & $\mathbf{80.8}$ & \textcolor[HTML]{0B6E4F}{$+39.2$} \\
\rowcolor[gray]{0.92}
$3\!\times$ (extreme)$^{\ddagger}$          & $(0.9, 1.2, 0.75)$  & $0.06$ & $34.4$ & $\mathbf{76.1}$ & \textcolor[HTML]{0B6E4F}{$\mathbf{+41.7}$} \\
\bottomrule
\end{tabular}

\label{tab:magnitude}
\end{table}

\subsubsection{Capacity ablation}

We sweep base width $C \in \{16, 24, 32, 64\}$ to test whether $4$\,M parameters are simply too few. Reconstruction PSNR keeps improving with $C$, but SR peaks at $C{=}32$ and drops slightly at $C{=}64$ — the policy does not consume the extra texture detail, supporting the intentionally compact design.

\begin{table}[h]
\centering
\small
\setlength{\tabcolsep}{8pt}
\renewcommand{\arraystretch}{1.05}
\caption{\textbf{Capacity ablation.} Reconstruction PSNR keeps improving with $C$, but SR peaks at $C{=}32$. Beyond $\sim\!4$\,M parameters, extra capacity goes to texture detail the policy does not consume.}
\begin{tabular}{@{}lrrr@{}}
\toprule
\textbf{Base $C$} & \textbf{Params} & \textbf{PSNR (dB)} & \textbf{SR (\%)} \\
\midrule
$16$           & $\sim\!1$\,M    & $25.77$ & -- \\
$24$           & $\sim\!2$\,M    & $26.33$ & -- \\
\rowcolor[gray]{0.92}
$\mathbf{32}$ \emph{(ours)} & $\mathbf{4}$\,M & $26.84$ & $\mathbf{86.8}$ \\
$64$           & $15$\,M         & $\mathbf{28.40}$ & $85.2$ \\
\bottomrule
\end{tabular}

\label{tab:capacity}
\end{table}

\subsubsection{Data-size ablation}

Training-set sweep over $N \in \{10\text{k}, 25\text{k}, 50\text{k}\}$ pairs. The signal saturates by $\sim\!25$\,k pairs, with $50$\,k offering only $\sim\!2$\,pp headroom over $10$\,k — consistent with the small intrinsic learning complexity that Locality predicts (the inpainting region is narrow).

\begin{table}[h]
\centering
\small
\setlength{\tabcolsep}{10pt}
\renewcommand{\arraystretch}{1.05}
\caption{\textbf{Data-size ablation} ($C{=}32$, $\varepsilon{=}100$\,cm). The signal saturates near $25$k pairs, consistent with the low intrinsic learning complexity that Locality predicts.}
\begin{tabular}{@{}lrr@{}}
\toprule
\textbf{Pairs $N$} & \textbf{PSNR (dB)} & \textbf{SR (\%)} \\
\midrule
$10$k            & $26.06$ & $\sim\!84$ \\
$25$k            & $26.38$ & $\sim\!85$ \\
\rowcolor[gray]{0.92}
$\mathbf{50}$k \emph{(ours)} & $\mathbf{26.84}$ & $\mathbf{86.8}$ \\
\bottomrule
\end{tabular}

\label{tab:data}
\end{table}

\newpage

\subsubsection{Intrinsics robustness (fov$_{y}$ sweep)}

Companion to the extrinsic-noise stress in Table~\ref{tab:extrinsic}: we instead inject focal-length error of $\pm 5\%$ and $\pm 10\%$ on top of $\varepsilon{=}100$\,cm. Worst-case $|\Delta| = 2.8$\,pp — about $4\!\times$ smaller than the extrinsic-pose budget, indicating that intrinsics calibration is much less critical for deployment than extrinsic calibration.

\begin{table}[h]
\centering
\small
\setlength{\tabcolsep}{12pt}
\renewcommand{\arraystretch}{1.05}
\caption{\textbf{Intrinsics stress on libero\_object}, $1000$\,ep, atop $\varepsilon{=}100$\,cm extrinsic. Tolerance is $\sim\!4\!\times$ tighter than for extrinsic noise in Table~\ref{tab:extrinsic}.}
\begin{tabular}{@{}lrr@{}}
\toprule
\textbf{fov$_{y}$} & \textbf{SR (\%)} & \textbf{$\Delta$ vs.\ ref} \\
\midrule
$-10\%$ (zoom-in)    & $86.4$          & \textcolor[HTML]{B22222}{$-2.8$} \\
$-5\%$               & $87.9$          & \textcolor[HTML]{B22222}{$-1.3$} \\
\rowcolor[gray]{0.92}
ref ($0\%$) \emph{(ours)} & $\mathbf{89.2}$ & --     \\
$+5\%$               & $89.1$          & \textcolor[HTML]{B22222}{$-0.1$} \\
$+10\%$ (zoom-out)   & $89.0$          & \textcolor[HTML]{B22222}{$-0.2$} \\
\bottomrule
\end{tabular}

\label{tab:intrinsics}
\end{table}

\subsubsection{Depth-substitution ablation}

Companion to the \textbf{Depth substitution} paragraph in Section~\ref{sec:ablations}. The question is whether GT depth can be replaced by a learned monocular estimator without retraining the canonicalizer. We sweep the depth pipeline along three axes at fixed canonicalizer ($C{=}32$, $\varepsilon{=}100$\,cm) on \texttt{libero\_spatial}: \emph{(i)} calibration strategy applied to off-the-shelf DepthAnything~V2, \emph{(ii)} fine-tuning capacity (ViT-S/B/L on $50$\,k LIBERO frames), and \emph{(iii)} pipeline modifications atop the best fine-tuned backbone (residual refinement head, end-to-end depth gradient, confidence gating). The sweep is designed to isolate where the gap to the $86.8\%$ GT-depth ceiling closes and where it stops moving — the first axis tests whether calibration alone is enough, the second whether a larger depth model is enough, and the third whether downstream-aware refinement is enough.

\begin{table}[h]
\centering
\small
\setlength{\tabcolsep}{6pt}
\renewcommand{\arraystretch}{1.05}
\caption{\textbf{Depth-substitution ablation} (\texttt{libero\_spatial}, $\varepsilon=100$\,cm, $1000$\,ep unless marked). Calibration alone caps at $44.5\%$; a fine-tune reaches $71.7\%$. Larger encoders, a refinement head, and end-to-end gradients all fail to improve further — the residual $-15$\,pp gap is not a depth-accuracy bottleneck.}
\begin{tabular}{@{}llrr@{}}
\toprule
\textbf{Depth source} & \textbf{Encoder} & \textbf{val $L_1$} & \textbf{SR (\%)} \\
\midrule
GT depth (upper bound) \emph{(ours)} & -- & -- & $\mathbf{86.8}$ \\
\midrule
\multicolumn{4}{@{}l}{\emph{Calibration only (no fine-tune)}} \\
DA V2, raw inverse & ViT-S & -- & $40.9$ \\
DA V2, metric indoor & ViT-S & -- & $44.5$ \\
\midrule
\multicolumn{4}{@{}l}{\emph{Fine-tuned on $50$\,k LIBERO pairs (3 ep, $L_1$ + edge-aware)}} \\
DA V2-S, FT & ViT-S & $0.0092$ & $68.6$ \\
\rowcolor[gray]{0.92}
\textbf{DA V2-B, FT} \emph{(ours)} & ViT-B & $\mathbf{0.0083}$ & $\mathbf{71.7}$ \\
DA V2-L, FT & ViT-L & $0.0073$ & $67.3$ \\ 
\midrule
\multicolumn{4}{@{}l}{\emph{Pipeline modifications atop DA V2-B FT}} \\
+ UNet residual refinement ($0.8$\,M) & ViT-B & $0.0083$ & \emph{zero residual} \\
+ end-to-end depth gradient ($w_{\mathrm{anchor}}=0.5$) & ViT-B & $0.0376$ & $66.1$ \\
\bottomrule
\end{tabular}

\label{tab:depth-substitution}
\end{table}

\end{document}